\documentclass[pdflatex,sn-mathphys-num]{sn-jnl}
\usepackage{float}

\usepackage{graphicx}%
\usepackage{multirow}%
\usepackage{amsmath,amssymb,amsfonts}%
\usepackage{amsthm}%
\usepackage{mathrsfs}%
\usepackage[title]{appendix}%
\usepackage{xcolor}%
\usepackage{textcomp}%
\usepackage{manyfoot}%
\usepackage{natbib}
\usepackage{booktabs}%
\usepackage{algorithm}%
\usepackage{algorithmicx}%
\usepackage{algpseudocode}%
\usepackage{listings}%
\usepackage[switch]{lineno}

\theoremstyle{thmstyleone}%
\theoremstyle{thmstyletwo}%

\theoremstyle{thmstylethree}%
\begin{document}

\title[How can reasoning capability empower the AI copilot robot in endoscopic surgery]{How can reasoning capability empower the AI copilot robot in endoscopic surgery}


\author[1]{\fnm{Guankun} \sur{Wang}}\email{gkwang@link.cuhk.edu.hk}
\equalcont{These authors contributed equally to this work.}

\author[1]{\fnm{Long} \sur{Bai}}\email{b.long@link.cuhk.edu.hk}
\equalcont{These authors contributed equally to this work.}

\author*[1]{\fnm{Hongliang} \sur{Ren}}\email{hlren@ee.cuhk.edu.hk}

\affil[1]{\orgdiv{Department of Electronic Engineering}, \orgname{The Chinese University of Hong Kong}, \state{Hong Kong}, \country{China}}

\abstract{
Reasoning capability has significantly advanced complex logical inference and robotic decision-making in general domains. However, its potential in the Artificial Intelligence (AI) copilot robot—particularly implemented based on the Vision-Language-Action (VLA) model—remains unexplored in endoscopic surgery. Effective reasoning should enable AI copilot robots to integrate multimodal cues, interpret surgical intent, and infer hidden tissue dynamics, thereby alleviating intraoperative uncertainty and cognitive burden on surgeons. Properly implemented, reasoning-driven autonomy can transform AI copilot robots from reactive executors into cognitive collaborators, enhancing precision, safety, and sustainability in clinical practice.}

\maketitle


\section{Introduction: AI Copilot Robot for Endoscopic Surgery}

Minimally Invasive Surgery (MIS) aims to reach targets through small access paths and to limit tissue injury. It can use small incisions or natural orifices, with visualization provided via endoscopic systems or from image-guided modalities such as fluoroscopy, Computed Tomography (CT), Magnetic Resonance Imaging (MRI), or Ultrasound (US). Conventional endoscopic surgery is a major part of MIS and includes rigid laparoscopy and flexible therapeutic endoscopy. 
Compared with open surgery, these methods can reduce incision size and perioperative complications. Yet conventional endoscopy has important limits: instrument angles and tip motion are restricted, counter-traction and two-hand coordination are hard, and operators face ergonomic strain. The view is often 2D from a single camera, and camera stability depends on manual control, which can cause tremor and changes in the field of view~\citep{haidegger2022robot}. These factors motivate robotic-assisted systems that seek to increase dexterity, stabilize vision, and reduce cognitive and physical load in endoscopic surgery.

Robot-assisted endoscopic surgery presents a transformative advancement in improving surgical precision, reducing patient trauma, and facilitating faster postoperative recovery~\citep{maier2017surgical}. To enhance the intelligence of robotic systems, VLA models have emerged as a promising solution, significantly advancing robotic automation in general domains~\citep{zheng2025universal, kim2025fine}. These models, built upon the Multimodal Large Language Models, are trained on large-scale, real-world robotic datasets, using visual inputs from cameras alongside textual task instructions to generate action outputs directly. However, deploying these technologies into the robot system to construct the AI copilot robot for endoscopic surgery is still challenging due to the complexity of manipulating deformable soft tissues \textit{in vivo}, such as cutting, tearing, and stretching~\citep{schmidgall2024general}. 

Specifically, AI copilot refers to a task-level/supervised autonomy assistant with Level of Autonomy (LoA) 2-3 (refer to~\citep{haidegger2019autonomy}), which can generate and monitor options and execute bounded low-level maneuvers under surgeon supervision, rather than a high-level autonomous system. During surgical operations, soft tissues often undergo unpredictable deformations in shape and position in response to external forces. This dynamic nature necessitates real-time monitoring of tissue deformation feedback to precisely adjust motion trajectories and applied forces, ensuring safety and efficacy during surgical procedures. 
Moreover, the inter-patient variability of soft tissues further increases the complexity of surgical operations, requiring the robot to exhibit strong adaptability to handle unexpected intraoperative scenarios. 
Typical examples include: sudden submucosal bleeding that requires rapid bleeding point localization and coordinated hemostasis, loss of traction or tearing of the mucosal flap that necessitates countertraction vector re-planning and dissection path adjustment~\citep{cui2022robotics, wang2025copesd, shao2024think}.
Therefore, serving as the cognitive engine of the AI copilot robot, VLA models must possess acute surgical environment awareness along with the ability to integrate contextual information for deep reasoning, akin to an experienced surgeon, to effectively handle complex surgical tasks.

\section{AI Copilot Definition and Autonomy Levels}

We use "AI copilot" to denote a surgical robotic assistant operating at LoA 2-3 by following the definition in~\citep{haidegger2019autonomy}, namely, task-level to supervised autonomy. Consistent with the standardization document of medical robot autonomy IEC/TR 60601-4-1, we frame autonomy along four Degree-of-Autonomy (DoA) functions—Generate, Execute, Monitor, and Select—and instantiate them with a reasoning-enabled VLA model that infers low-level motion goals and multi-instrument coordination options from high-level intent and visual context (Generate), maintains situation awareness of tissue state, tools, and uncertainty (Monitor), and carries out bounded low-level maneuvers under explicit safety constraints with immediate surgeon override available at all times (Execute), while leaving task-level option selection to the surgeon and permitting only limited low-level option selection by the system within pre-approved safety envelopes (Select). Under this definition, the surgeon remains in- or on-the-loop with ultimate authority (including emergency stop and override) and retains responsibility for task-level decisions. Our objective is not to advance toward LoA 4-5, but to strengthen LoA 2-3 capabilities—particularly option generation and monitoring alongside safe low-level execution—through explicit reasoning to improve precision, enhance safety, and reduce cognitive load.

\section{Multimodal Sensing and Uncertainty-Aware Fusion for the AI Copilot Robot}

A key capability of an AI copilot robot for endoscopic surgery is to combine information into a surgeon-supervised model, using a VLA abstraction to align modalities, resolve conflicts, and weight sources by uncertainty when suggesting actions~\citep{maier2017surgical, cui2022robotics, zhang2024ai}. Beyond video, preoperative CT/MRI provide priors; intraoperative imaging such as Endoscopic Ultrasound (EUS) and Optical Coherence Tomography (OCT) reveal subsurface vessel likelihood; sensors such as shape sensing, Electromagnetic (EM) tracking, and force proxies stabilize localization. Rather than treating video as truth, the copilot keeps a probabilistic map that persists through occlusions and anticipates subsurface risk: priors guide margins, adjunct maps set depth and avoid high vessel-probability regions, and pose stabilization buffers loss. Reasoning is pivotal to conditionally re-weight sources as conditions change (e.g., defer to priors during smoke/blood occlusion, elevate EUS/OCT influence when subsurface danger is suspected) and to propagate input uncertainty into conservative action constraints and interpretable surgeon feedback. In endoscopic resection, it fuses lesion extent with deformation and shape sensing to suggest traction-dissection; in bronchoscopy, it combines airway maps with cues to lower bleeding risk~\citep{cui2022robotics, zhang2024ai}. Realization needs real-time deformable pre-/intraoperative alignment, uncertainty-aware fusion, and benchmarks linking fusion quality to safety-critical outcomes, while preserving surgeon authority and override.

\section{Reasoning in VLA Models for Endoscopic Surgery}

\begin{figure*}[ht]
    \centering
    \includegraphics[width=\textwidth]{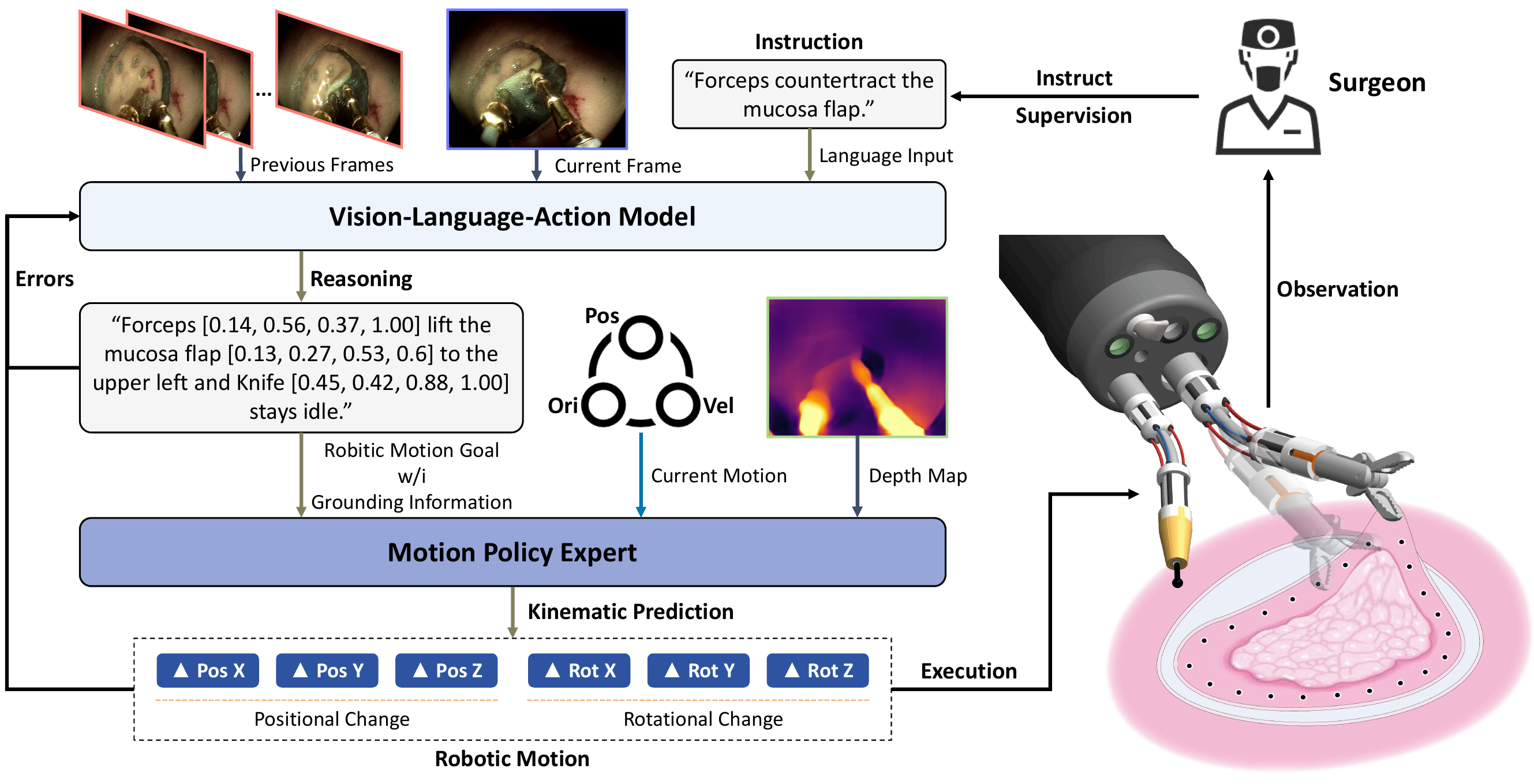}
    \caption{Proposed reasoning-driven architecture for the AI copilot robot in endoscopic surgery. The first VLA model performs reasoning based on high-level language instructions and surgical video frames, generating low-level instrument motion interactions as the Robotic Motion Goal aligns with grounding information. The reasoning output, combined with additional multimodal information, is then converted into kinematic changes by the second VLA model, which serves as a motion policy expert. The following abbreviations are used throughout the figure: Pos (Position), Ori (Orientation), Vel (Velocity).}
    \label{fig}
\end{figure*}

Reasoning in the VLA model refers to the capability to perform multi-step logical deduction and integrate diverse information streams before generating action outputs, as shown in Figure~\ref{fig}. Early work from institutions such as OpenAI and Google DeepMind introduced chain-of-thought prompting, which significantly enhanced model performance in areas such as mathematics, programming, and scientific reasoning. Recently, DeepSeek-R1~\citep{guo2025deepseek} has further advanced the field by refining chain-of-thought methods and integrating reinforcement learning within a multi-stage training paradigm. The primary motivation for incorporating reasoning capability is to enable models to go beyond simple pattern matching, achieving a deeper understanding of the underlying problems and generating self-explanatory solutions~\citep{mon2025embodied}. 
This advanced capability not only improves generalization across diverse domains and enhances adaptability in dynamic endoscopic surgery but also bolsters the robustness of models when confronting soft tissue manipulation.

Specifically, reasoning capability empowers the AI copilot robot in several key ways. First, reasoning enables the robot to flexibly interpret user commands and visual information from the surgical environment. The work of Co-Pilot of Endoscopic Submucosal Dissection~\citep{wang2025copesd} has demonstrated that VLA models can reason low-level motion instructions and grounding information based on simple language instruction prompts and visual features in endoscopic surgery. Such low-level goal instructions align more closely with automated execution and can provide more accurate guidance for following action generation. For instance, when a surgeon issues a command to ``\textit{Forceps countertracts the mucosa flap.}", the model assesses the deformation state of the tissue, the position of instruments and the optimal dissection area to generate the motion instruction and the direction of each instrument in detail. This means that the VLA model is not simply executing pre-defined commands but is actively analyzing the context to generate precise countertraction-dissection coordination. Similarly, during hemostasis, reasoning enables the robot to identify bleeding points under challenging visual conditions (e.g., occlusion by blood or smoke) and coordinate instrument actions—such as stabilizing tissue with one forceps while precisely applying cautery with another—to achieve rapid and safe bleeding control. It is important to note that AI copilot robots are not yet capable of continuously executing an entire surgical workflow. Consequently, their current role is confined to specific, well-defined subtasks (e.g., traction, hemostasis). In these contexts, the surgeon provides task-level intent cues, which are interpreted by the VLA model with intraoperative variations to produce accurate low-level motions. Such a division of labor ensures clinical feasibility and reduces the cognitive load on the surgeon.

Second, reasoning capability can bolster the VLA model's proficiency in managing the intricate interactions among multiple surgical instruments. Unlike general robotic tasks where grasping is the primary action, endoscopic surgery involves the collaboration of multiple instruments with distinct functions, such as cauterization, injection, and grasping, etc~\citep{gao2024transendoscopic, cui2022robotics}. Instrument collaboration and action diversity require precise coordination to ensure efficient and safe execution of surgical procedures. Reasoning shall play a crucial role in comprehensively understanding the dynamic relationship between the instruments and the tissue, which facilitates the VLA model to effectively coordinate robot actions to achieve seamless, synchronized movements. In particular, during endoscopic submucosal dissection, the forceps stabilize and manipulate the tissue while the knife executes controlled dissection in perfect synchronization. This coordinated control, underpinned by reasoning, allows the AI copilot robot to prevent potential conflicts between instruments and adjust its motions in real-time~\citep{firoozi2025foundation}.

Third, the integration of the chain-of-thought reasoning process empowers VLA models for both anticipatory planning and uncertainty-aware decision-making. Given the non-linear mechanical properties of soft tissues and inter-patient variability, predicting the effects of surgical maneuvers is inherently complex. Reliable uncertainty quantification is therefore crucial, as even minor miscalculations can lead to significant clinical consequences. Chain-of-thought reasoning allows the model to iteratively simulate multiple future states, evaluate the consequences of various actions, and refine its strategy in real-time to optimize dissection precision while mitigating the risk of collateral tissue damage~\citep{shao2024think}. When deployed within the AI copilot robot, this reasoning capability facilitates tissue response anticipation, allowing the system to respond dynamically to tissue deformation, unexpected bleeding, and other intraoperative uncertainties.


Finally, reasoning enables continuous learning and adaptation of VLA models during the deployment through multiple surgical procedures~\citep{kim2025srt}. By incorporating feedback loops that monitor both the immediate and downstream effects of its actions, the model can refine its internal models continuously. This self-correcting capability is analogous to the reflective thinking exhibited by experienced surgeons, who continuously adjust their technique based on tactile and visual cues. The inclusion of reinforcement learning strategies, where the model is incentivized to explore and optimize its decision-making, can further accelerate this adaptive process, driving the system toward higher levels of autonomy and efficacy.

\section{Sustainability Considerations for AI Copilot Endoscopic Surgery}

Embedding sustainability as an explicit objective in AI copilot endoscopic surgery is essential to avoid shifting burdens to public systems~\citep{guenat2022meeting}. In line with evidence from the United Nations Sustainable Development Goals (SDGs) that robotics can advance many targets yet hinder others through environmental footprint and inequities~\citep{haidegger2023robotics}, we argue that reasoning in AI copilot robots is a direct lever for sustainable deployment. Multi-step reasoning can plan bimanual maneuvers that has the potential to minimize tool exchanges, smoke/irrigation, and tissue trauma—shortening operative time, anesthesia, and disposable use while reducing re-interventions. Uncertainty-aware “when-to-think” policies trigger deeper analysis only under risk or ambiguity, avoiding unnecessary computation. Structure-aware reasoning uses priors and causal task knowledge to learn and transfer from fewer annotated cases, reducing data collection and retraining cycles. By co-optimizing clinical reward with resource costs (e.g., consumables), the copilot can prefer equally safe but less resource-intensive plans and apply sustainability constraints (such as favoring reusable instruments when equivalent). Operating robustly with simpler sensor suites further eases infrastructure demands and supports equitable adoption. We recommend reporting per-case time/resource/energy alongside outcomes to make sustainability auditable.

\section{Conclusions and Future Directions}
The integration of reasoning capability into VLA models represents a transformative step toward realizing robust AI copilot robot systems for endoscopic surgery. By enabling multi-step logical deduction, contextual awareness, and dynamic adaptation, reasoning empowers AI copilot robots to achieve surgeon-like decision-making under uncertainty. However, challenges persist in scaling these capabilities to the computational and temporal constraints of real-world surgical procedures. A central direction for future work is to optimize inference pipelines through lightweight architectures, model pruning, and edge hardware acceleration, which is crucial to achieving sub-second response times. While such optimization improves efficiency, it also introduces trade-offs between computational compactness and reasoning reliability, potentially compromising decision robustness in complex surgical contexts. This limitation highlights the importance of establishing rigorous reliability assurance frameworks—including uncertainty-calibrated reasoning and surgeon-in-the-loop supervision—to ensure that reasoning-driven AI copilot robots remain safe and dependable even under constrained computational budgets.

For deployment, the integration of reasoning-driven models into existing surgical robotic platforms requires seamless communication with control systems, including kinematic interfaces and haptic feedback modules. Efforts to establish interoperability frameworks and  Application Programming Interfaces (APIs) will accelerate clinical translation. Moreover, adaptive explainability strategies should be developed to balance real-time responsiveness with transparency. For instance, while core reasoning remains optimized for speed, detailed reasoning traces could be selectively disclosed in low-confidence situations or upon surgeon request, thereby ensuring oversight without compromising intraoperative real-time performance. Beyond deployment, cross-institutional collaboration will be critical to establishing standardized benchmarks for evaluating reasoning fidelity and safety-critical failure modes. Notably, our discussion on integrating reasoning capabilities into endoscopic surgery is equally applicable to other surgeries, where similar challenges in tissue manipulation and multi-instrument coordination exist. Ultimately, by embedding reasoning capabilities within VLA models, AI copilot robots are promised to be transformed from reactive executors into cognitive collaborators that not only assist surgeons but also enhance procedural precision, safety, and the long-term sustainability of surgical practice.

\section{Acknowledgements}
This work was supported by HK RGC, Collaborative Research Fund (CRF C4026-21GF), General Research Fund (GRF 14203323, GRF 14216022, and GRF 14211420), NSFC/RGC Joint Research Scheme N\_CUHK420/22, Shenzhen-Hong Kong-Macau Technology Research Programme (Type C) STIC Grant 202108233000303.

\section{Author contributions statement}
Guankun Wang: Conceptualization, Investigation, Writing – original draft, Figure creation. Long Bai: Conceptualization, Investigation, Writing – original draft. Hongliang Ren: Conceptualization, Supervision, Writing – original draft, Figure creation.

\section{Competing interests}
The authors declare no competing interests.

\bibliography{sn-bibliography}

\end{document}